\documentclass[nohyperref]{article}

\usepackage{microtype}
\usepackage{graphicx}
\usepackage{subfigure}
\usepackage{booktabs} % for professional tables

\usepackage{hyperref}

\usepackage[accepted]{icml2023}

\usepackage{amsmath}
\usepackage{amssymb}
\usepackage{mathtools}
\usepackage{amsthm}

\usepackage[capitalize,noabbrev]{cleveref}

\theoremstyle{plain}
\newtheorem{theorem}{Theorem}[section]

\newtheorem{lemma}[theorem]{Lemma}

\theoremstyle{definition}
\newtheorem{definition}[theorem]{Definition}
\newtheorem{assumption}[theorem]{Assumption}
\theoremstyle{remark}

\usepackage[textsize=tiny]{todonotes}

\icmltitlerunning{A Robust Hypothesis Test for Tree Ensemble Pruning}

\begin{document}

\twocolumn[
\icmltitle{A Robust Hypothesis Test for Tree Ensemble Pruning}



\icmlsetsymbol{equal}{*}

\begin{icmlauthorlist}
\icmlauthor{Daniel de Marchi}{yyy}
\icmlauthor{Matthew Welch}{xxx}
\icmlauthor{Michael Kosorok}{yyy}
\end{icmlauthorlist}

\icmlaffiliation{yyy}{Department of Biostatistics, University of North Carolina at Chapel Hill}
\icmlaffiliation{xxx}{Colby College}

\icmlkeywords{Machine Learning, ICML}

\vskip 0.3in
]



%\printAffiliationsAndNotice{} % leave blank if no need to mention equal contribution

\begin{abstract}
Gradient boosted decision trees are some of the most popular algorithms in applied machine learning. They are a flexible and powerful tool that can robustly fit to any tabular dataset in a scalable and computationally efficient way. One of the most critical parameters to tune when fitting these models are the various penalty terms used to distinguish signal from noise in the current model. These penalties are effective in practice, but are lacking in robust theoretical justifications. In this paper we develop and present a novel theoretically justified hypothesis test of split quality for gradient boosted tree ensembles and demonstrate that using this method instead of the common penalty terms leads to a significant reduction in out of sample loss. Additionally, this method provides a theoretically well-justified stopping condition for the tree growing algorithm. We also present several innovative extensions to the method, opening the door for a wide variety of novel tree pruning algorithms. 
\end{abstract}

\section{Problem Introduction}

Gradient boosted tree algorithms are some of the most popular models in applied machine learning, as they are more flexible than linear models and typically more robust than neural networks out of sample \cite{goldbloom2016}. They operate by optimizing for the minimum information-theoretic divergence (such as entropy or Gini impurity) between the predictions of tree leaves and the true target values, where the model output is the conditional expectation of the outcome of a number of binary splits of the input variable. The model is grown in a progressive fashion, where individual trees are fit to fix the errors of the prior ensemble. This is in contrast to the other popular tree ensemble algorithm, random forests, where each tree is intended to be an effective estimator \cite{banfield2006comparison}. This iterative ensembling tends to outperform the bagging ensembling method of random forests. Both methods significantly outperform single trees \cite{goldbloom2016,krauss2017deep,zhang2021prediction}. 

Gradient boosted trees have a number of software advantages as well. The most used packages were originally developed by high performance computing specialists, so they are also adapted to modern computing hardware, and can be trained on both GPUs and distributed computing systems. Modern gradient boosted decision tree packages typically work like second-order Taylor approximators of the conditional expected value of a target variable, rapidly converging to an effective solution \cite{chen2016xgboost}. 

The theory and methodology of gradient boosted trees is underserved relative to how frequently these models are used, though that is not to say innovative work is not being done. There are some significant efforts to move the individual tree construction away from greedy optimization and towards a global optimization paradigm, which delivers numerous improvements \cite{gabidolla2022pushing}. An investigation has also been done into pruning decision trees with genetic algorithms, and there has been some work into new ways to evaluate splits under difficult conditions such as concept drift \cite{mijwil2021utilizing, wang2021evolving}. Confidence intervals for trees are somewhat well-studied, though there is constant innovation to further develop the techniques \cite{duan2020ngboost, mentch2016quantifying}. However, asymptotically valid hypothesis tests to evaluate tree splits have not been developed to any significant extent. To our knowledge, the only statistically developed formal hypothesis test for tree splits is the Chi-Square Automatic Interaction Detection (CHAID) method. This evaluates splits by running a Chi-Square test between the current data in the parent node and the data within the two child nodes after the prospective split. Denote $y_{E}$ to be the expected $y$ purity if the split does not improve the purity of the parent node, and $y_{O}$ be the actual purity after the split. The p-values of the $\chi^2$ statistics above and below the split are evaluated for each possible split \cite{kass1980exploratory}. 

\[SplitQualityCHAID = \frac{\sqrt{(y_{O} - y_{E})^2}}{y_{E}}\]

This method runs afoul of several issues. It is asymptotically very similar to the gradient, entropy, or impurity formulations, making the differences in splits attained with this method and other methods only marginally different. It is also fundamentally difficult to perform so many tests in a robust way. Especially when testing thousands of splits across dozens of covariates at every split evaluation step, it is difficult to justify running so many hypothesis tests, even when a significance correction is applied. Such corrections are not a cure-all, and it is better to do a more modest amount of tests than simply significance-correct thousands of ad hoc tests \cite{perneger1998s}. Finally, it only works for classification tasks.

Instead of formal hypothesis tests, decision tree ensembles commonly use four parameters to control overfitting : a learning rate, a minimum gain (whether in information or loss reduction) to produce a split, an $L_1$ split penalty, and an $L_2$ split penalty. Especially after an effective set of parameters have been found via a validation set or cross validation, some combination of these typically produces a robust model \cite{natekin2013gradient}. However, the theoretical justification for these parameters is lacking. 

$L_1$ and $L_2$ parameters, which originated in regression, are well-grounded for that type of model. However, trees are nonparametric models and do not have coefficients to shrink. Instead, they shrink the split quality metrics towards zero. These $L_1$ and $L_2$ parameters were "hacked" into trees because of their strength in other models. Similarly, setting a minimum gain value is intuitive but also lacks solid theoretical grounding without any sort of consideration for the distribution of the gain, which is impossible to derive explicitly in finite data sets. Raising or lowering a learning rate is also intuitive, as a progressive tree weighting scheme would make sense to deliver better results. Again though, the justification is lacking \cite{chen2015xgboost}.

To address these issues we developed a hypothesis test that evaluates the split quality after a tree has been fit. This hypothesis test can both be used to prune trees and to set an optimal stopping condition. Applying this method to several real-world datasets led to large improvements in out of sample loss. We specifically used gradient boosted trees because the tree pruning procedure developed can produce very homogeneous trees if single trees are fit independently, making it unsuitable for a random forest approach. There are also significant advantages to having a progressively recomputed gradient and Hessian, which will become clear later. While this method would also theoretically work for pruning a single decision tree, these are rarely used, so we do not consider that for this paper. Therefore we focus for the remainder of the paper on gradient boosting algorithms.

\subsection{Gradient Boosting Details}

Here, we rederive some useful properties of gradient boosting and use them to lead into the hypothesis test procedure. Consider a univariate $Y$ and a possibly multidimensional $X$, made up of $n$ paired examples $(x_1, y_1)...(x_n, y_n)$. Assume we are trying to find some function $t(X) = Y$, where $t$ is a gradient boosted tree algorithm. Assume $X$ has $j$ dimensions, denoting the columns of $X$ to be $X_{1}...X_{j}$. Finally, denote some unknown measure space for each variable, $(\Omega_X, \mathcal{F}_X, f_X), (\Omega_Y, \mathcal{F}_Y, f_Y)$, as well as a joint distribution space $(\Omega_{X, Y}, \mathcal{F}_{X, Y}, f_{X, Y})$. The individual trees of the estimating function $t$ can be represented as $t_k$. We can then represent the current prediction as the sum of the $K$ trees.

\[\hat{Y} = \sum_{k=1}^K t_k(X), t_k \in \mathcal{T}\]

Here, $\mathcal{T}$ is the space of possible trees given the current hyperparameter settings and the characteristics of $X$ and $Y$. Assume then that we are minimizing a regularized loss objective between the current tree ensemble and our target variable, $Y$.

\[\mathcal{L}(Y, \hat{Y}) = \sum_{i=1}^n l(y_i, \hat{y_i}) + \sum_{k=1}^K \mathcal{R}(t_k)\]

Here, $l$ is the loss function between $Y$ and $\hat{Y}$, and $\mathcal{R}$ is a regularization function on the trees. $\mathcal{L}$ is the overall loss we are trying to minimize. Then, at the k'th step, we can formulate the objective.

\[\mathcal{L}(Y, \hat{Y})^{(k)} = \sum_{i=1}^n l(y_i, \hat{y_i}^{(k-1)} + t_k(x_i)) + \mathcal{R}(t_k)\]

$\mathcal{L}(Y, \hat{Y})^{(k)}$ denotes the loss at the k'th iteration, and $\hat{y_i}^{(k-1)} = \sum_{i=1}^{k-1} t_i(X)$.

Gradient boosting approximates this by using a second-order Taylor approximation. 

\[\mathcal{L}(Y, \hat{Y})^{(k)} \approx \sum_{i=1}^n l(y_i, \hat{y_i}^{(k-1)} + g_i t_k(x_i) + \frac{1}{2} h_i t^2_k(x_i)) + \mathcal{R}(t_k)\]

Here, $g_i, h_i$ are the first and second derivatives of the loss with respect to the estimate at step $k-1$. For example, for mean squared error (the most common regression loss) the first derivative $g$ will be the vector of differences, $Y_1 - \hat{Y_1}...Y_n - \hat{Y_n}$ and the second derivative $h$ will be a constant vector of ones across the space. These terms are denoted as the gradient and Hessian by the XGBoost package, which we preserve for the rest of this work. This matrix of gradients and the Hessian matrix are critical to formulate our hypothesis test. We selected XGBoost to use as the formulation of gradient boosting for the other popular packages, lightGBM and Catboost, are similar \cite{ke2017lightgbm, dorogush2018catboost}

\section{Hypothesis Test Setup}

To clarify this section, we define some additional properties of a split under the modern gradient boosting definition. The cover is the number of points that the split affects in the training set. The gain, which calculates the improvement in the loss function, is defined by the equation below. From here out, "gain" denotes the unregularized gain function. While penalty terms for gain exist, we use the hypothesis test to penalize it going forward.

\[Gain = \frac{(G_L)^2}{H_L} + \frac{(G_R)^2}{H_R} - \frac{(G_L + G_R)^2}{H_L+H_R}\]

Here, $G_L$ is the sum of the gradients below the split, and $G_R$ is the sum of the gradients greater than or equal to the split. Similarly, $H_L$ is the sum of the Hessian below the split, and $H_R$ is the sum of the Hessian greater than or equal to the split. 

\begin{assumption}
\label{ass:one}
The residuals, gradient and Hessian of the current step of a gradient boosting algorithm are random variables.
\end{assumption}

\begin{lemma}
\label{thm:two}
Correctly parameterized gradient boosted decision trees are universal approximators.
\end{lemma}

Denote an individual split in $t_j$ to be $I(X_{V_{j, k}} \leq S_{j, k})$, where $1 \leq j \leq J$ is the index of the tree, $1 \leq k \leq K$ is the index of the split node in that tree, $V_{j, k}$ is the column of $X$ that the split is operating on, and $S_{j, k}$ is the value of the split boundary. The outcome of the split is then a Bernoulli random variable, which we will denote as $I_{j, k}$.

We can apply the simple function approximation theorem. A simple function is any function of the form $\sum_{1}^n a_k I(X \in A_k)$ for a finite collection of disjoint measurable sets $A_1...A_k$ on a measure space $\Lambda$, and real-valued coefficients $a_k$. By this definition, the random variables $I_{j, k}$ are clearly simple functions approximating the true distribution of $E(Y|X=x)$. By the simple function approximation theorem, we know that if we allow the gradient boosting algorithm to construct an arbitrary number of trees, and have an arbitrary amount of data, $E(Y|I_{1, 1}, I_{1, 2}...I_{J, K}) \to E(Y|X)$ pointwise. This means that unregularized gradient boosting trees are unbiased estimators of the true conditional distribution of $Y|X$, assuming that the number of trees is allowed to grow arbitrarily \cite{breiman2017classification}. This then implies gradient boosted decision trees are universal approximators, because the distribution of $Y|X$ was left arbitrary, meaning the trees could learn to approximate any distribution.

\begin{definition}
\label{def:one}
A perfectly fit tree ensemble is a model such that $E(Y|X) = E(Y|I_{1, 1}, I_{1, 2}...I_{J, K})$. By (2.2), such a model always exists.
\end{definition}

\begin{theorem}
\label{thm:two}
A new split on tree ensemble, even a perfectly fit ensemble (as in definition 2.3) will still have unregularized gain greater than zero.
\end{theorem}

Assume that the tree model is correctly fit to the true distribution, i.e. $E(Y|X) = E(Y|I_{1, 1}, I_{1, 2}...I_{J, K})$ for some fit tree model. Assume we were assessing whether or not to add a $J+1$th tree, but that to do so would be overfitting the true conditional distribution, and $E(Y|X)$ may no longer equal $E(Y|I_{1, 1}, I_{1, 2}...I_{J, K}, I_{J+1, 1})$.

Let $G$ be the current sum of gradients of the decision tree and $H$ be the sum of the Hessian. Then define four sub-variables, $G_R, G_L, H_L, H_R$, the sums of the gradient and Hessian above and below some prospective split. The loss function used to generate $G, H$ is necessarily a convex function on real values, and trivially that implies that the gradient $G$ will be real-valued, and the Hessian values will be greater than zero by properties of convex functions.

Then the Cauchy-Schwarz inequality (specifically see Titu's lemma to the theorem) immediately states that $\frac{(G_L + G_R)^2}{H_L+H_R} \leq \frac{(G_L)^2}{H_L} + \frac{(G_R)^2}{H_R}$, which means the gain of the split will be greater than zero, even though it harms the model. This is why regularization of trees is necessary - even a perfect tree ensemble will continue to train if unregularized.

It might also have been sensible to conclude the gradient boosting algorithm would deliver nonzero gain if it was not a universal approximator. Non-universal approximators can go in circles, optimizing for one factor at the expense of another back and forth, never actually improving. Universal approximators do not have this issue. Therefore, no matter how perfectly our model is currently fit, a nonzero gain on a subsequent split is inevitable.

\begin{lemma}
\label{thm:three}
If a random variable $V$ contains no information about the variable of interest $Y$, or the covariates used for prediction $X$, the expected gradient of splits on $V$ is zero. The gain will still be greater than zero.
\end{lemma}

If a random variable $V$ contains no information about the variable of interest $Y$, or about the covariates used to predict $Y$, $X$, then by definition of information $I(V, Y) = D_{KL}(P_{VY}||P_V P_Y) = 0$, where $D_{KL}$ is the Kullback-Leibler divergence function. Similarly, $I(V, X) = 0$ as well.

Since gradient boosting is a conditional expectation optimizer, if $I(V, Y) = 0, E(Y|V) = E(Y)$, and therefore we would not expect any splits on $V$ to improve our predictions. Similarly, $E(X|V) = E(X)$ if $I(X, V)$ = 0. Therefore, informational content about $V$ neither serves to inform us about $Y$, nor is there a mediated relationship where $V$ informs us in some way about $X$, which could in turn inform us about $Y$. Excluding this possible mediated relationship is essential to the development of the theorem. For instance, consider $X$ was a copy of $Y$ plus some noise, and $V$ was the noise variable. Then, even though $I(V, Y) = 0$, we could say that $X-V = Y$, and so $E(Y|X, V) \neq E(Y|X)$ and $\mathcal{L}(Y, E(Y|X, V)) \leq \mathcal{L}(Y, E(Y|X))$ for some loss function $\mathcal{L}$.

If the mutual information between $V$ and both $X$ and $Y$ is zero, then since $E(Y|V, X) = E(Y|X)$, the expected gradient must be zero as $\mathcal{L}(E(Y|X), Y) = \mathcal{L}(E(Y|X, V), Y)$, so the slope of the optimization function with respect to $V$ is zero. 

As proved in theorem 2.4 however, the gain on an uninformative variable will still be greater than zero, even though the expected gradient with respect to this variable is zero. We refer to sampling from the possible gains of an information-free variable as in this lemma to be sampling from the "null" gain distribution, i.e. the possible gains of useless variables. This leads into our hypothesis test.

\section{The Hypothesis Test Procedure}

Assume we have a candidate split, where we are trying to evaluate if the gain of this split is due to signal or noise. Theorem 2.4 tells us that no matter what, this split will have positive gain. To assess if the candidate split is a genuine improvement in the model, we can compare its gain to the gain of a sample from the "null" gain distribution.

For the cover $C$ of our candidate split, generate a random sample of size $C$ from our existing vector of gradients $G$, Hessian $H$, and target variable $Y$, denoted $G_C, H_C, Y_C$. Independently, generate a size $C$ random sample from a permutation of $Y$, denoted $R_Y$. A split on $R_Y$ is the purely uninformative split we want to compare our candidate split to. $R_Y$ clearly fits the definition of an information-less variable for lemma 2.5, as it has no relation to either $X$ or $Y$ due to being a sample of a random permutation of $Y$. Using our prior theorems, it will be a split with gradient zero with respect to the optimization problem, but gain greater than zero.
 
\subsection{The Test}

Pick a random split on $R_Y$, denoted $S_{R_Y}$. Separate $G_C, H_C$ into $G_{L_C}, G_{R_C}, H_{L_C}, H_{R_C}$ above and below the split, then calculate the gain using the formula given earlier. Compare this gain to the gain of the candidate split. If the gain of this randomized split is greater than the gain of the candidate split, then the split is unlikely to be genuine as it produced a lower gain than a split on a completely uninformative variable. If it is less, then it is better than a completely random split. 

If we denote the split being tested to be $I_{j_o, k_p}$, and the sequence of prior splits $I_{j_1, k_1}...I_{j_o, k_{p-1}}$, the null hypothesis of this test is that $E(\mathcal{L}(E(Y|X) - E(Y|I_{j_1, k_1}...I_{j_o, k_p}))) \geq E(\mathcal{L}(E(Y|X) - E(Y|I_{j_1, k_1}...I_{j_o, k_{p-1}})))$, i.e. the conditional loss is unchanged or worsened by adding or removing the candidate split and it is equally as good as the split on $R_Y$. If this is the case, then the probability that the gain of the random split exceeds that of the split being tested will be 0.5. This is because under the null, both variables produce gains from the same unknown distribution of "null" gains, positive gains due purely to accidentally significant partitions of the gradient and Hessian. Regardless of the variance, the probability any two that the one draw of a random variable will exceed a second draw of the same variable is 0.5.

This is a binary hypothesis test with probability of success under the null 0.5, so we can test the hypothesis that the variables are unrelated to any type 1 error rate of the form $\alpha = \frac{1}{2^k}$ for some integer $k$. This is akin to repeated coin tossing. Redraw $R_Y$ $k$ times and get $k$ draws from the "null" gain distribution. If the gain of the candidate split exceeds all $k$ null draws, then the candidate split is genuine with type 1 error rate $\frac{1}{2^k}$. Otherwise, we accept the null, that the candidate split's gain is no better than a draw from the "null" gain distribution.

\begin{lemma}
\label{thm:three}
Hypothesis test pruning of individual splits also provides a hypothesis test to terminate growing additional trees, with type 1 error rate proportional to the type 1 error rate of the pruning of individual splits.
\end{lemma}

Assume we are pruning a tree using the hypothesis test above, so we prune a split if it does not pass the hypothesis test to level $\alpha$. Then, assuming the tree is of depth $D$, that implies we performed $\sum_{i=1}^{D-1} 2^i = T$ such tests. If we define the stopping condition to be whether or not the entire tree is pruned, then we are implicitly performing a test with type 1 error rate $\alpha^T$ of whether or not growing additional trees is necessary. We use this as the stopping condition for our future results.

\subsection{The Correlation Variant}

With hypothesis tests for small or noisy datasets, it is common to want to be able to increase the conservativeness of the test. For our test, this can easily be done by increasing the strength of the relationship between our permuted $R_Y$ and the true $Y$. The correlation between a randomly shuffled $Y$ and $R_Y$ is 0 since they have no relationship, which is our original formulation. However, if we only permute every other element then $R_Y$ will have a linear relationship to $Y$ with expected correlation 0.5. By shuffling a fixed proportion of the $Y$ and leaving the rest, we can generate any arbitrary linear correlation $\rho$ in expectation. This is because the union set of a set of $K_Y$ with correlation $1$ of size $k$ with a set of $J_Y$ with correlation $0$ of size $j$ will have total correlation to $Y$ of $\frac{k}{k+j} = \rho$. This is proved in lemma 3.2 at the end of this section.

This allows us to perform our hypothesis test with any possible relationship strength. In fact, an unpermuted $R_Y$ with correlation 1 to the true $Y$ represents the maximum possible relationship strength, and will deliver the highest possible gain for a single split. For instance, if we want to test splits relative to a random variable with correlation 0.01, we could generate an $R_Y$ where every hundredth observation is unpermuted. This allows us to easily compare the gains of a variable which may have some complex, high dimensional relationship to our target $Y$ onto the simple relationship of linear correlation. If we set $\rho$ to 0.5, and the gain of our candidate split is greater than the gain on $R_Y$, we can be extremely confident that the candidate split is a genuine improvement. A random variable with correlation 0.5 is an extremely useful variable for most statistical modeling tasks. In contrast, if we find that the split only delivers gain similar to an $R_Y$ with correlation 0.001 to the true $Y$, we would probably be extremely dubious that there is a genuine relationship.

 This also makes the $\rho$ correlation parameter to be a useful hyperparameter to tune in cross-validation as a margin of safety to the hypothesis test above. If the data size is limited, or if there is significant noise, it may be useful to build a more conservative test to prevent false discovery and add additional robustness to the hypothesis test, at the cost of potentially disregarding weaker relationships and increasing the model bias. It also does not increase compute time, in fact increasing $\rho$ will marginally decrease compute time as we no longer have to permute those indices.
 
\begin{lemma}
\label{lem:usefullemma}
A random variable $V$ that is a associated with another variable $Y$ such that a random sample of size $n$ of $(V, Y)$ will have $k$ examples of $V$ exactly equal to $Y$ and $j$ examples equal to a random shuffling of a draw from the distribution of $y$, such that $k+j = n$, will have expected correlation $\frac{k}{k+j}$ to $Y$.
\end{lemma}

Consider $Corr(V, Y)$ = $\frac{E((V - \mu_V)(Y - \mu_Y))}{\sigma_V \sigma_Y}$. Decompose $V$ to two random variables,$V_k$ and $V_j$, where $V_k$ is the perfectly matching sample and $V_j$ is the randomly shuffled sample.  We know that $Var(Y)$ = $Var(V)$ = $Var(V_k)$ = $Var(V_j)$. We also trivially know that $Corr(Y, V_k) = 1 = \frac{E((V_k - \mu_{V_k})(Y - \mu_Y))}{\sigma_{V_k} \sigma_Y}$, and that $Corr(Y, V_j) = 0$ = $\frac{E((V_j - \mu_{V_j})(Y - \mu_Y))}{\sigma_{V_j} \sigma_Y}$. Finally, we can say $Corr(Y, V) = \frac{P(V \in V_j)E((V_j - \mu_{V_j})(Y - \mu_Y)) + P(V \in V_k)E((V_k - \mu_{V_k})(Y - \mu_Y))}{\sigma_{V} \sigma_Y} = \frac{P(V \in V_k)E((V_k - \mu_{V_k})(Y - \mu_Y))}{\sigma_{V} \sigma_Y} = P(V \in V_k) * 1 = \frac{k}{k+j}$.

\subsection{Computational Complexity}

Generating random indices is an $O(n)$ process, and accessing these indices from a vector is also $O(n)$. Therefore generating our random sample of the gradient and Hessian of size equal to the size of the split $c$ is an $O(c)$ process. Generating the $R_Y$ has similar computational complexity. For generating and accessing an $R_Y$ with correlation to $Y$ of $\rho$, selecting random indices is an $O(n-n\rho)$ process, and accessing the corresponding $Y$ is an $O(n)$ process. Overall, the worst case complexity of the procedure is $O(4n)$ when performing this procedure on the first split, where the size of the split equals the number of examples in the dataset.

Contrast this to actually evaluating the split itself. Most gradient boosted tree algorithms use a histogram split to speed up the process. Generating a histogram with $k$ buckets is an $O(kn)$ process, and evaluating the $k$ potential splits of the histogram on a dataset of size $n$ requires $kn$ operations. This totals to $2(k+1)n$ operations. Doing this for each variable, where there are $j$ variables, brings the final time complexity to fit a decision tree split to $O((k+1)*j*n)$

Therefore the complexity of our split evaluation is relatively low in comparison to the total complexity of a gradient boosted tree fitting algorithm. It adds approximately a worst-case $4n$ operations to what is currently a worst-case $2(k+1)*j*n$ process for each split. In most gradient boosted tree algorithms, $k$ is set quite large, with 256 as the default value in XGBoost and LightGBM.

In the worst case, we quintuple computational complexity if there is only one binary variable, $X$, as then there is only one split to evaluate. If there are multiple $X_{i}$ which are all continuous, and thus exceeds the histogram count, then we multiply computational complexity by $\frac{2(k+1)*j*n+4n}{2(k+1)*j*n}$, where $k$ is the number of buckets in the histogram. This increase will on average tend to be quite small since $k$ is usually fairly large, and the typical datasets used in gradient boosting contain many features.

We did not achieve these bounds in practice. Our coding was done in python, which is orders of magnitude slower than C++, the language XGBoost is written in. We also had to recompute many terms XGBoost uses in training, as the XGBoost backend is not exposed to us as modifiable functions. Our current implementation is significantly slower than XGBoost overall.

\section{Methods and Data}

To compare the two methods, we searched a large parameter grid using 5-fold cross validation to assess the effectiveness of different hyperparameter combinations on multiple datasets (described later). For standard XGBoost, we tuned a parameter grid on the learning rate, tree depth, gamma (minimum gain for a split), lambda (L2 penalty), and the number of estimators. We tuned these across the grid of all possible hyperparameter combinations, utilizing exhaustive grid search. Typically this was around 1200 tested combinations, except in the two Bitcoin experiments where we searched 1440 combinations. The best hyperparameter combination (by area under the receiver operator characteristic curve for the classifiers, and mean absolute error for the regression models) was saved and trained on the full training set. For our method, we searched across a grid of the maximum depth, learning rate, the per-split $\alpha$ of our hypothesis test, and the correlation of the random $R_Y$ to the true $Y$ as described in the subsection (2.1). Due to computational constraints, we only searched significantly fewer hyperparameter combinations than for vanilla XGBoost, typically 72 combinations (except the Bitcoin experiments where it was 96) or only 6\% (6.667\% for the Bitcoin experiments) of the typical number in vanilla XGBoost. We did not run additional methods such as CatBoost or LightGBM because they do not deliver significant performance gains relative to XGBoost and have nearly identical gradient boosting formulations. For specifics on the hyperparameter ranges, as well as download links to the datasets, please see the attached code.

We pulled five datasets from Kaggle to run a total of six experiments. These datasets are publicly available and relatively well-studied, making them perfect for reproducible investigations. We selected three regression datasets and two classification datasets to showcase the performance of the hypothesis test procedure on multiple loss functions. For evaluation metrics, we used mean absolute error on the regression datasets, and the ROC AUC for the binary classifiers.

When there were nominal categorical features in the dataset, we one-hot encoded them. For more details on one hot encoding, please see \cite{agresti2012categorical}. Features that could not be easily coerced to numeric features were dropped. Values were otherwise left un-normalized. The prediction task was kept the same for our work and the Kaggle events the datasets were associated with. The datasets are summarized in the table below.

\begin{table}[h]
\caption{Dataset Summaries}
\label{sample-table}
\vskip 0.15in
\begin{center}
\begin{small}
\begin{sc}
\begin{tabular}{lcccr}
\toprule
Name & Train Size & Test Size & Objective \\
\midrule
Bitcoin A & 2242 & 748 & Regression\\
Bitcoin B & 2243 & 748 & Regression\\
Water & 2185 & 1092 & Classification\\
Shipping & 3666 & 7334 & Classification\\
Airfoil & 1127 & 376 & Regression\\
Avocado & 3650 & 14600 & Regression\\
\bottomrule
\end{tabular}
\end{sc}
\end{small}
\end{center}
\vskip -0.1in
\end{table}

The Airfoil dataset is a NASA data set, obtained from a series of aerodynamic and acoustic tests of two and three-dimensional airfoil blade sections conducted in an an echoic wind tunnel. It is a regression task dataset where the goal is the predict the scaled sound pressure level in decibels. The Shipping dataset is a dataset of whether or not packages reached customers on time, with covariates about both how the package was shipped and characteristics of the customer. The Avocado dataset is a dataset of Avocado prices and sales in several markets. The Water dataset is a dataset of water potability and various characteristics of water samples. Finally, both Bitcoin datasets are price and volume information starting in 2013.

The two Bitcoin datasets require additional explanation. Both are actually the same dataset featurized different ways. The data contains covariates for time, volume, and the opening, closing, high and low prices. To normalize the data (since the price of Bitcoin was extremely contingent upon the time) we converted all of these values to percent changes from the prior day.

In the first experiment (Bitcoin A) we attempt to predict the price change for the next day using values of the prior day, a fairly straightforward way to use the pricing data. In the second dataset (Bitcoin B) we do not attempt to predict future prices. Instead, we try and predict the intraday percent change of the current day, which is equal to the percent change of the open of the current day versus the close. This can be expressed as a perfect function of inputs, and a theoretically optimal model would have zero loss because a perfect function exists. However, there is a multicolinearity issue, where the time, volume, high, and low are all correlated with the target as shown in Bitcoin A. We theorized that our hypothesis test will learn this relationship better than vanilla gradient boosting, and view it as an interesting added experiment. This was inspired by a discussion linked on the XGBoost documentation on whether gradient boosting can learn addition \cite{chen2016xgboost}.

\section{Results}

We first summarize the in-sample loss below, as this showed several interesting trends in the data. We show the regression data first, where lower scores are better as models were evaluated with mean absolute error (MAE). When referring to our method, we refer to them as HTrees (hypothesis trees) to contrast with XGBoost.

\begin{table}[h]
\caption{In-Sample Results (Regression)}
\label{sample-table}
\vskip 0.15in
\begin{center}
\begin{small}
\begin{sc}
\begin{tabular}{lcccr}
\toprule
XGBoost & Min MAE & Mean MAE & Max MAE \\
\midrule
Bitcoin A & 0.0235 & 0.0271 & 0.0412\\
Bitcoin B & 2.6e-4 & 0.021 & 0.31\\
Airfoil & 0.914 & 1.925 & 3.821\\
Avocado & 0.064 & 0.13 & 0.16\\
\toprule
HTrees & Min MAE & Mean MAE & Max MAE \\
\midrule
Bitcoin A & 0.0238 & 0.0268 & 0.0307\\
Bitcoin B & 2.8e-4 & 6.4e-4 & 1.2e-3\\
Airfoil & 0.89 & 1.01 & 1.17\\
Avocado & 0.064 & 0.068 & 0.077\\
\bottomrule
\end{tabular}
\end{sc}
\end{small}
\end{center}
\vskip -0.1in
\end{table}

For the classification datasets, we used ROC AUC (ROC) as a metric. Here, a higher metric is better.

\begin{table}[h]
\caption{In-Sample Results (Classification)}
\label{sample-table}
\vskip 0.15in
\begin{center}
\begin{small}
\begin{sc}
\begin{tabular}{lcccr}
\toprule
XGBoost & Min ROC & Mean ROC & Max ROC \\
\midrule
Water & 0.5 & 0.566 & 0.681\\
Shipping & 0.693 & 0.742 & 0.772\\
\toprule
HTrees & Min ROC & Mean ROC & Max ROC \\
\midrule
Water & 0.617 & 0.651 & 0.682\\
Shipping & 0.727 & 0.748 & 0.769\\
\bottomrule
\end{tabular}
\end{sc}
\end{small}
\end{center}
\vskip -0.1in
\end{table}

One of the more surprising outcomes of using hypothesis-tested trees is that they appeared to be less sensitive to hyperparameter perturbations. While the vanilla XGBoost models are searching a significantly larger grid of parameters, it is still somewhat surprising that the difference between the minimum and maximum performance is so much larger in XGBoost. This is an extremely useful property to have, as it implies that when the hyperparameters are misconfigured the performance reduction will be smaller for hypothesis-driven trees than in classical gradient boosting. We now show results on the out-of-sample test sets.

\begin{table}[h]
\caption{Out of Sample Comparison (Regression)}
\label{sample-table}
\vskip 0.15in
\begin{center}
\begin{small}
\begin{sc}
\begin{tabular}{lcccr}
\toprule
Dataset & XGBoost MAE & HTrees MAE \\
\midrule
Bitcoin A & 0.0258 & 0.0261\\
Bitcoin B & 9.2e-3 & 3.6e-4\\
Airfoil & 4.687 & 3.8\\
Avocado & 0.42 & 0.37\\
\bottomrule
\end{tabular}
\end{sc}
\end{small}
\end{center}
\vskip -0.1in
\end{table}

\begin{table}[h]
\caption{Out of Sample Comparison (Classification)}
\label{sample-table}
\vskip 0.15in
\begin{center}
\begin{small}
\begin{sc}
\begin{tabular}{lcccr}
\toprule
Dataset & XGBoost ROC & HTrees ROC \\
\midrule
Water & 0.624 & 0.641\\
Shipping & 0.729 & 0.729\\
\bottomrule
\end{tabular}
\end{sc}
\end{small}
\end{center}
\vskip -0.1in
\end{table}

From this table we can see that while there was a significant improvement in mean performance, the distribution is extremely uneven. In the Bitcoin A and Shipping datasets, there is either no improvement or a small performance reduction. The Water dataset shows a moderate performance increase. The Avocado, Airfoil, and Bitcoin B experiments show substantial performance gains. Note again that this is despite the fact that the hypothesis-driven trees had dramatically fewer hyperparameter combinations tested, and are only using one method to penalize the trees as opposed to several. Their performance was also better correlated to their performance in cross-validation, with the ratio between the out of sample performance and in-sample performance much closer to 1 for the HTrees.

\begin{figure}[h]
    \centering
    \includegraphics[width=8cm]{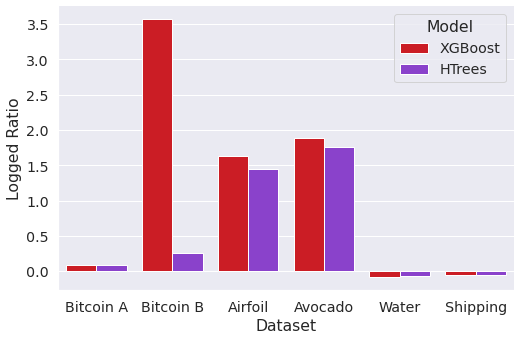}
    \caption{Logged Ratio Between Out of Sample and In-Sample Metrics}
\end{figure}

There are a few important takeaways. The first is that the larger the fluctuation between the maximum and minimum performance during cross validation, the more likely the hypothesis trees were to outperform. We can see that the gap between the best and worst hyperparameter combinations was very large in Avocado, Airfoil, and Bitcoin B, and relatively small in Water, Bitcoin A and Shipping. This implies that strong hyperparameter dependence is an indicator of the hypothesis-driven tree pruning method outperforming.

\begin{figure}[h]
    \centering
    \includegraphics[width=8cm]{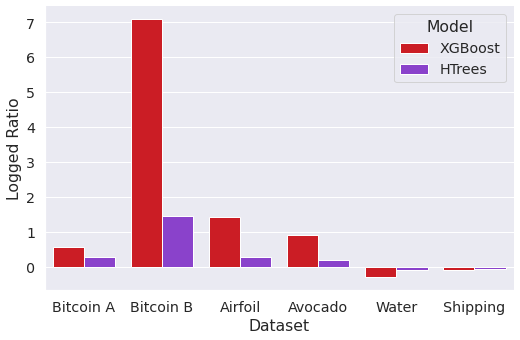}
    \caption{Logged Ratio Between Best and Worst Cross Validation Metrics for Hypothesis Trees vs. XGBoost}
\end{figure}

The second is that, even in relatively bad cases, hypothesis-driven trees perform about equally to ordinary gradient boosting. The worst performance difference is on Bitcoin A, where the gap in mean absolute error was 0.0003, or only 1\%. This is a relatively insubstantial gap that would likely disappear if more hyperparameter combinations were tested.

However, this outperformance would not necessarily be predictable just by examining the relative performances of the best models in cross validation. The XGBoost had slightly better cross validation metrics on both Bitcoin experiments and the Shipping dataset, and slightly lower cross validation metrics on the other three datasets. While Bitcoin A and Shipping are the two best performances for XGBoost, Bitcoin B is by far the worst. This is an issue, as it implies that there may not be an easy way to tell when XGBoost may perform better. Cross validation that uses both the hypothesis test and conventional means of regularization may be optimal for real-world problems for this reason.

\section{Discussion}

This work makes a powerful and novel argument for increased statistical theory in the gradient boosting literature. Theoretical properties of gradient boosting are relatively neglected compared to the robust development of the theory of linear models and the explosion of research on deep neural networks (DNNs). 

The primary benefit is increased out of sample performance in worst-case scenarios. Depending on the data and problem, there are differing expectations about out of sample performance. Real-world data generating distributions are often highly unstable, and more fragile relationships between covariates are the most susceptible to rapidly become misleading if the data-generating distribution changes. Having increased confidence in the true relationship between split indicators and the target variable appears to dramatically increase robustness to these shifts.

Finally, this work makes a small improvement in the "black box" nature of machine learning models. Hypothesis testing exists to prevent faulty relationships from making it into the final analysis. A model capable of being hypothesis tested is inherently a more explainable model because with this method, you can specify a relationship and explain how the trees were assessed against that relationship. L1 and L2 parameters to regularize the gain may work just as well, but saying that each split in the tree has been tested such that it is better than a random variable with correlation $\rho$ with probability $\alpha$ is an inherently more theoretically sound description, mapping black box models onto elementary statistical relationships that can be explained to almost anyone.

\section{Future Work}

This work has several notable areas for future work. First, but extremely significantly, it is not integrated with XGBoost. The authors wrote efficient code, however this means that our method is slower than it would be if it were fully incorporated into the XGBoost package, which is written in a high performance C++ backend. This limits the current scalability of this method, and would make it much slower than XGBoost on extremely large datasets or when extensively testing a large number of hyperparameter combinations on a single dataset. This also may have degraded our performance relative to XGBoost since we searched significantly fewer hyperparameter sets for our method than we did for XGBoost.

In addition, our current hypothesis test is not particularly customizable. In particular, there has been much research done on false discovery rate control, hypothesis testing for high dimensional data, and corrections for multiple testing that could be incorporated into our method, allowing for a wide variety of alternative user-specified testing methods. Depending on the problem, a more customizable testing function could deliver huge improvements.

Allowing some way to adjust for the false discovery rate is therefore likely the most significant next theoretical step \cite{benjamini1995controlling, he2016component}. As trees learn, the value of $\alpha$ should initially be high, then decrease for later trees. We expect that assuming that the covariates are not totally useless for predicting the target, it will be easier to find significant splits earlier when the models are simplistic as opposed to later, when the models are more sophisticated.

Finally, this method could incorporate a correction for information bias. It has been demonstrated that tree split metrics bias towards splitting on features with more levels \cite{deng2011bias}. The primary issue is that the intrinsic information of a variable rises with the number of unique values. All tree split metrics are proportional to the information of the feature, with higher-information features producing (on average) higher gain on unregularized splits. This would be a fairly novel step for tree learning packages, as to the knowledge of the authors such a correction is not implemented in XGBoost or LightGBM, the two most popular packages.

Overall, while we believe this work is a meaningful contribution to the gradient boosting literature, it is an important first step towards more robust and more theoretically grounded tree pruning algorithms.

\bibliography{example_paper}
\bibliographystyle{icml2023}

\end{document}